# A Three-Pillar Governance Framework for Safe and Trustworthy AI Agents

Edward C. Cheng
Management Science & Engineering
Stanford University
Stanford, California,
United States
echeng04@stanford.edu

Jeshua Cheng
AI Laboratory
Inquiryon, Inc.
South San Francisco, California,
United States
jeshua.cheng@inquiryon.com

Alice Siu
Deliberative Democracy Lab
Stanford University
Stanford, California,
United States
asiu@stanford.edu

***Abstract*—AI agents capable of autonomous decision-making introduce significant challenges related to safety, alignment, accountability, and human oversight. Existing Human-in-the-Loop (HITL) and responsible AI approaches provide important foundations, but many remain fragmented across system design, governance, and operational deployment. This paper presents a Three-Pillar (3P) model and framework for developing safe and trustworthy AI agents through transparency, accountability, and trustworthiness. The framework defines an evolutionary path in which increasing agent autonomy is introduced through progressive validation, structured human oversight, and auditable operational controls. We further describe how the model supports agent lifecycle monitoring, decision traceability, escalation management, and human-agent collaboration within enterprise agentic environments. By integrating oversight principles with operational mechanisms, the proposed framework provides a practical foundation for deploying AI agents that remain aligned with human intent while enabling controlled progression toward higher autonomy.**

## I. Introduction

AI agents represent a significant evolution in generative AI systems, extending beyond content generation toward autonomous reasoning, planning, and task execution. Unlike traditional conversational systems, AI agents can perform actions on behalf of users, including information retrieval, transaction execution, workflow coordination, and interaction with external tools and services. While these capabilities create substantial opportunities for productivity and automation, they also introduce new risks related to hallucination, bias, incomplete context, unsafe actions, and misalignment between agent behavior and human intent.

Recent research has increasingly emphasized the importance of Human-in-the-Loop (HITL) approaches, collaborative AI systems, and uncertainty-aware agent architectures to improve controllability, safety, and alignment in real-world deployments [1]–[6]. These efforts collectively suggest that fully autonomous agents should not be deployed through immediate end-to-end automation, but instead through progressively increased autonomy in which human oversight remains integrated throughout the evolution of the system.

However, existing approaches remain fragmented across model alignment, reinforcement learning, uncertainty estimation, human oversight, and institutional governance. Many focus on improving individual aspects of safety and controllability without providing a unified operational framework for deploying and managing autonomous agents within enterprise environments.

This paper presents a Three-Pillar model framework for safe and trustworthy AI Agents. The framework defines three foundational conditions for managed autonomy: *transparency, accountability,* and *trustworthiness*. Together, these pillars establish an operational environment in which agent behavior remains observable, decisions remain attributable, and trustworthiness is reinforced through human oversight implemented through Human-in-the-Loop (HITL) policy and escalation mechanisms. The framework further defines an evolutionary path toward autonomy in which increasing agent capability is introduced through progressive validation, structured oversight, and continuous governance.

The contributions of this paper are threefold:

1) We synthesize prior work in Human-in-the-Loop AI, collaborative agents, uncertainty-aware systems, and AI governance into a unified framework for safe agent operation.
2) We introduce a Three-Pillar (3P) model framework that operationalizes transparency, accountability, and trustworthiness within agentic environments.
3) We illustrate how the framework supports progressive autonomy through structured human oversight, lifecycle monitoring, decision traceability, and operational mechanisms.

The remainder of this paper is organized as follows. Section II reviews related work in Human-in-the-Loop AI, collaborative agent systems, and AI orchestration. Section III discusses the evolutionary path toward autonomous agents. Section IV presents the proposed Three-Pillar Framework. Section V illustrates the framework through a representative enterprise agent workflow. Section VI discusses future directions and concludes the paper.

## II. Related Work

### A. Human-in-the-Loop and Collaborative AI

Human-in-the-Loop (HITL) approaches have emerged as an important foundation for improving the safety, controllability, and alignment of AI systems. Early work established HITL as both a technical and organizational mechanism for integrating human expertise into AI development and operation. Zanzotto introduced Human-in-the-Loop Artificial Intelligence (HitAI) as a framework in which humans remain central to both knowledge creation and system oversight [2]. Subsequent surveys and taxonomies expanded this view by framing HITL as a collaborative paradigm that combines human cognition with machine scalability through iterative feedback and supervision [3], [4].

More recent work has extended HITL concepts into operational agent systems. Bellos and Siskind demonstrated that interactive AI guidance can improve task completion and long-term user performance in complex physical tasks [7]. Mozannar et al. introduced Magentic-UI, a Human-in-the-Loop interface for agentic systems that incorporates structured oversight mechanisms such as approval and verification workflows [8]. Together, these efforts illustrate a broader shift from fully automated AI toward collaborative human-agent systems in which oversight remains integrated into execution workflows.

### B. Uncertainty-Aware and Governed AI Systems

Recent research has further advanced safe autonomy through reinforcement learning, uncertainty estimation, and governance-oriented AI frameworks. Retzlaff et al. argued that reinforcement learning systems inherently depend on human feedback and identified explainability, trust calibration, and feedback quality as key requirements for safe deployment [9]. Ren et al. proposed uncertainty-aware planning mechanisms in which AI agents proactively request human assistance under low-confidence conditions, reducing unsafe autonomous behavior [10].

Beyond technical safeguards, researchers have increasingly emphasized institutional governance and democratic oversight of AI systems. Allen et al. proposed risk-management models centered on transparency, public accountability, and shared authority in AI decision-making [11], while Barroso and Mello highlighted the importance of human dignity, accountability, and democratic governance in large-scale AI deployment [12]. Natarajan et al. further argued that many systems traditionally described as Human-in-the-Loop are more accurately characterized as AI-assisted decision systems in which humans retain ultimate authority [13].

### C. Relationship to Existing Governance Frameworks

The proposed 3P framework is complementary to established AI governance and risk frameworks. NIST AI RMF provides organizational functions for governing, mapping, measuring, and managing AI risk [14]; ISO/IEC 42001 defines requirements for an organizational AI management system [15]; the EU AI Act establishes risk-based regulatory requirements including transparency, traceability, human oversight, and risk management [16]; and the OWASP Top 10 for LLM Applications identifies security risks and mitigations relevant to LLM and agentic applications [17].

These frameworks establish important governance objectives, requirements, and risk controls, but do not prescribe a unified runtime model for governing increasing agent autonomy. The 3P framework addresses this operational layer by organizing governance around transparency, accountability, and trustworthiness and implementing these principles through lifecycle monitoring, decision traceability, executable policies, HITL escalation, and progressive delegation of authority. It is therefore intended to complement existing standards by providing an agent-specific operational framework through which broader governance requirements can be implemented during execution.

### D. Limitations of Existing Approaches

Although prior work has significantly advanced Human-in-the-Loop AI, collaborative agent systems, uncertainty-aware planning, and AI orchestration, many approaches remain fragmented across technical, operational, and institutional dimensions. Existing frameworks often address specific aspects of safety and alignment without defining a unified operational environment for governing autonomous agents throughout their lifecycle.

In particular, limited attention has been given to how transparency, accountability, and trustworthiness can be operationalized together within continuously operating agentic systems. Furthermore, many approaches emphasize either human oversight or autonomous capability independently, without defining a structured evolutionary path through which autonomy can increase under measurable controlled and validation conditions.

These limitations motivate the need for an integrated management framework that combines operational transparency, decision accountability, and enforceable human oversight to support the safe progression of AI agent autonomy.

## III. Progressive Evolution Toward Autonomous Agents

### A. Controlled Reduction of Human-in-the-Loop

The development of autonomous AI agents requires a staged progression in which increasing autonomy is introduced through validation, monitoring, and governance. Similar to the evolution of autonomous driving systems, AI agents should transition incrementally from assistive behavior toward governed autonomy rather than immediate end-to-end automation [18], [19], [20]. This staged approach enables systems to improve through operational feedback while maintaining human oversight in high-risk or ambiguous situations.

HITL mechanisms play a central role in this progression by enabling humans to validate outputs, intervene during failures, and provide corrective feedback for system improvement. Prior research has shown that HITL improves interpretability, reliability, and trust calibration in collaborative AI systems [3], [4], [7], [9]. As agents demonstrate consistent and predictable behavior, human intervention can be reduced progressively under defined escalation policies.

This evolutionary approach is particularly important in trust-sensitive domains such as finance, healthcare, legal services, and enterprise HR operations, where unsafe autonomous behavior may produce significant operational or societal consequences. Progressive autonomy therefore requires not only advances in technical capability, but also governance mechanisms that support transparency, oversight, accountability, and risk management throughout the agent lifecycle.

### *B. Stages of Agent Autonomy*

The progression toward autonomous agents can be viewed as a sequence of operational stages that balance automation with human oversight:

*1)* **Assisted Agents**, where humans retain decision authority while AI systems provide recommendations or analysis.

*2)* **Collaborative Agents**, where humans and agents jointly participate in planning and execution workflows.

*3)* **Supervised Autonomy**, where agents execute constrained tasks independently under human review and escalation controls.

*4)* **Governed Autonomy**, where agents operate with higher levels of autonomy within transparent, auditable, and policy-governed operational boundaries.

Progression across these stages should be driven by demonstrated reliability, alignment, and operational safety rather than assumptions of readiness. This requires management frameworks capable of monitoring agent behavior, enforcing escalation policies, maintaining auditability, and supporting controlled delegation of authority as agent capability evolves.

## IV. Three-Pillar Governance Framework

### *A. Framework Overview*

The transition from assistive AI systems to autonomous agents requires operational environments that support observability, accountability, and controlled human oversight. Without such mechanisms, increasing autonomy may introduce unacceptable risks related to unsafe actions, misalignment, and loss of human control.

To address these challenges, this paper proposes a Three-Pillar (3P) Model consisting of transparency, accountability, and trustworthiness. The governance framework constructed from the 3P model defines the operational conditions required for deploying AI agents under governed and progressive autonomy, where increasing agent capability is introduced through structured oversight, policy enforcement, and successive validation.

The proposed framework is designed to support continuously operating agentic systems in enterprise environments, enabling lifecycle monitoring, decision traceability, escalation management, and HITL intervention throughout agent execution.

### *B. Transparency*

Transparency requires observability across the full lifecycle of an AI agent, including *initiation, execution, escalation, completion,* and *termination* states. In the 3P agentic environments, transparency enables humans and supervisory systems to monitor agent behavior, inspect decision processes, reconstruct execution history, and audit operational outcomes.

During initiation, the operating environment records the agent's objectives, constraints, permissions, configuration parameters, and associated governance policies. These records establish the operational context under which the agent is authorized to execute tasks [21].

During execution, the environment captures activity journals that record actions, tool usage, intermediate outputs, external interactions, state transitions, and HITL interventions. When escalation conditions are triggered, human approvals, overrides, and corrective actions are recorded as part of the execution trace.

Upon completion or termination, the system maintains an auditable record of the full execution lifecycle. At minimum, transparency requires:

*1)* lifecycle state transition records,

*2)* activity and execution logs,

*3)* decision and escalation records,

*4)* HITL interaction records.

These operational records support monitoring, compliance, post-execution analysis, and continuous improvement of agent behavior.

Fig. 1 illustrates the agent lifecycle states, execution transitions, and Human-in-the-Loop escalation points supported by the proposed 3P framework.

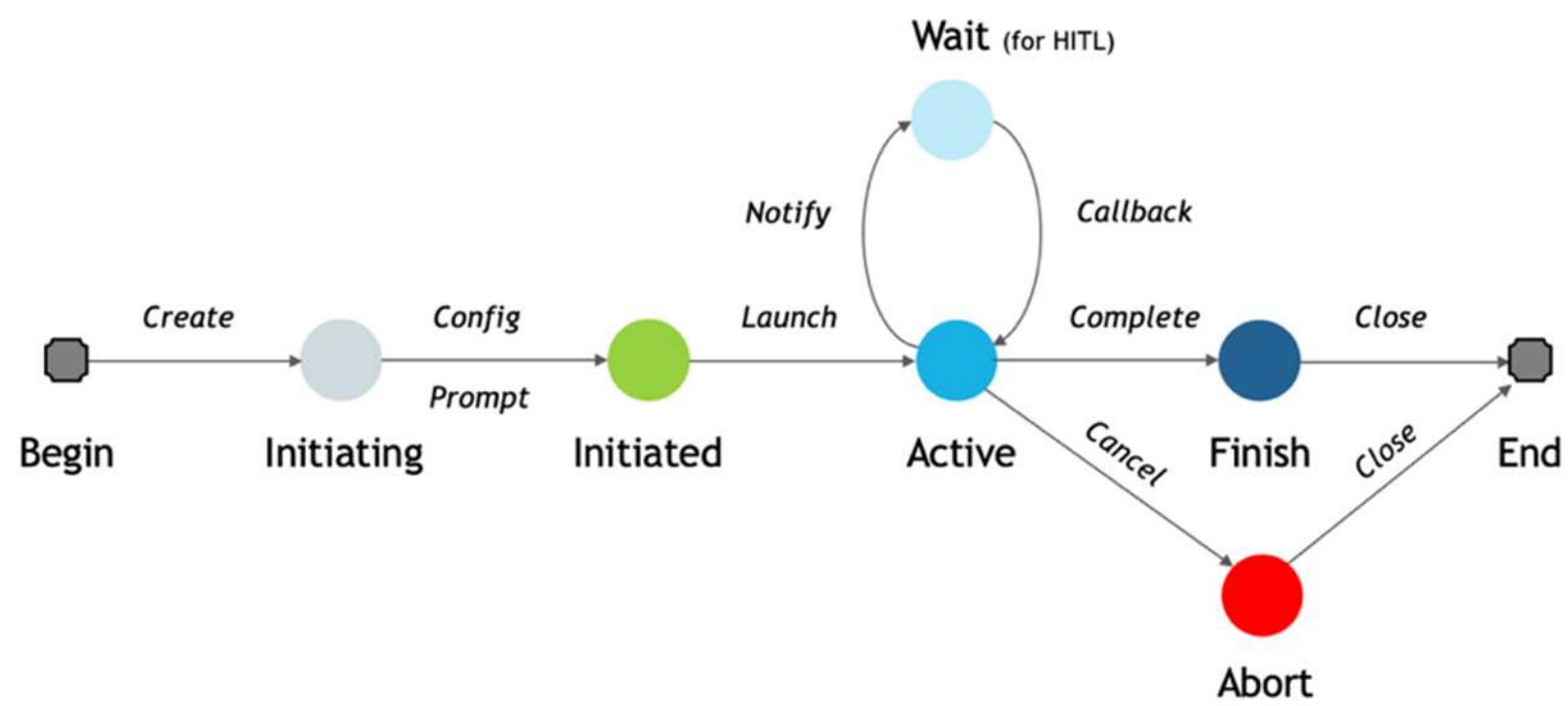

Fig. 1. Agent state transition diagram within the Three-Pillar governance framework.

### C. Accountability

Accountability requires that agent decisions, actions, and escalation events remain attributable and explainable throughout the execution lifecycle. In governed agentic environments, accountability enables organizations to reconstruct operational decisions, identify failure points, enforce policy compliance, and assign responsibility across human and AI participants.

To support accountability, the operating environment maintains decision journals that record execution context, reasoning traces, policy conditions, input sources, and resulting actions. Human approvals, overrides, and corrective interventions are also recorded to preserve traceability across Human-in-the-Loop (HITL) workflows.

These records support post-execution auditing, operational review, incident analysis, and iterative improvement of agent behavior. In high-impact domains, accountability mechanisms are particularly important for validating compliance with organizational policies, regulatory requirements, and safety constraints.

Accountability also supports operational correction and system improvement by enabling organizations to identify failure points, implement corrective actions, and reduce recurrence of unsafe behavior. These mechanisms strengthen the feedback loop required for managed autonomy and continuous refinement of agent behavior.

### D. Trustworthiness and Human Oversight

Trustworthiness requires that autonomous behavior remain subject to controlled delegation, policy enforcement, and risk-based human oversight. Rather than eliminating human involvement, governed autonomy introduces mechanisms that regulate when agents may operate independently and when escalation or intervention is required.

The framework supports HITL controls through configurable approval thresholds, escalation policies, fallback mechanisms, and supervisory intervention workflows. High-risk or low-confidence actions may require explicit human authorization, while lower-risk tasks may execute with reduced supervision under predefined operational constraints.

As agent performance improves, oversight requirements may be adjusted progressively based on validated operational behavior. However, all increases in autonomy remain subject to policy controls, monitoring, and auditability to ensure that operational safety and alignment are maintained.

### E. Progressive Governance of Autonomy

The proposed framework supports progressive autonomy through staged delegation of authority under continuous oversight and validation. Rather than treating autonomy as a binary property, the framework enables gradual transitions from assisted operation toward increasingly autonomous execution as agents demonstrate reliable performance within defined operational boundaries.

In early stages, human oversight remains central to execution and decision approval. As operational confidence increases, agents may assume greater responsibility under policy thresholds and escalation controls. Operational safeguards remain active throughout this progression, including monitoring, auditability, intervention policies, and rollback capabilities. Importantly, the three pillars of transparency, accountability, and human oversight is maintained across all stages of progressive autonomy, while direct human participation is progressively reduced as agent automation demonstrates reliable performance.

This approach enables organizations to balance operational efficiency with safety, accountability, and human control while supporting continuous adaptation of agent behavior in dynamic environments.

## V. Operational Use Case

### A. Enterprise Customer Services Email Agent

To illustrate the proposed framework, consider an enterprise email agent responsible for processing inbound customer requests, retrieving organizational information, generating draft responses, and coordinating follow-up actions. Such systems operate in environments where incorrect actions, incomplete context, or unauthorized decisions may introduce operational, legal, or reputational risks.

The agent workflow is composed of multiple operational stages, including request interpretation, information retrieval,

decision generation, escalation handling, and response execution. Throughout execution, oversight mechanisms remain active to ensure that autonomous actions are observable, attributable, and subject to human oversight.

The use case is an illustrative implementation of the proposed framework and is not presented as a production deployment or empirical evaluation.

### *B. Governance Through the Three Pillars*

Transparency and explainability are supported through lifecycle monitoring and activity journals that record agent actions, tool usage, execution states, escalation events, and HITL interactions. These records provide operational visibility and enable reconstruction of agent execution history.

Accountability is supported through decision traceability, policy enforcement records, escalation logs, and audit mechanisms that preserve attribution across both AI and human participants. These records support operational review, compliance analysis, and corrective improvement following execution failures or policy violations.

Trustworthiness is maintained through HITL approval workflows, escalation policies, and risk-based intervention mechanisms that regulate autonomous execution. High-risk actions may require explicit human approval, while lower-risk tasks may execute under predefined policy constraints.

With these three pillars maintained across all stages of progressive autonomy, agent actions and decisions remain observable, auditable, and subject to human oversight throughout the operational lifecycle.

Fig. 2 illustrates representative activity records and execution traces captured during operation of an enterprise email agent. Fig. 3 illustrates a Human-in-the-Loop approval workflow used to support human escalation and supervisory intervention. Together, these figures illustrate how transparency is supported through observable execution records, accountability through traceable decisions and actions, and trustworthiness through enforceable human oversight.

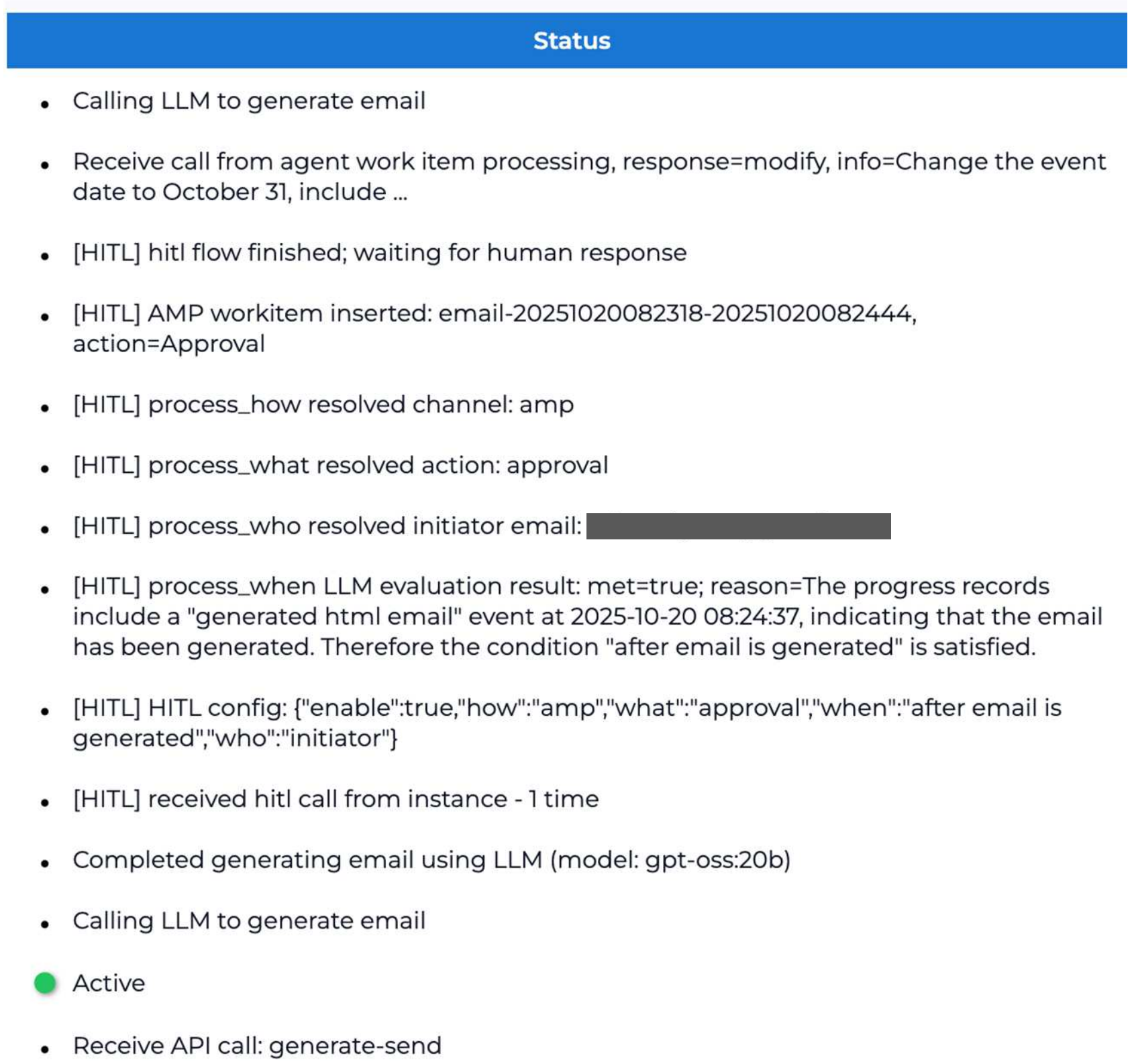

Fig. 2. Agent activity records captured during execution of an enterprise email agent.

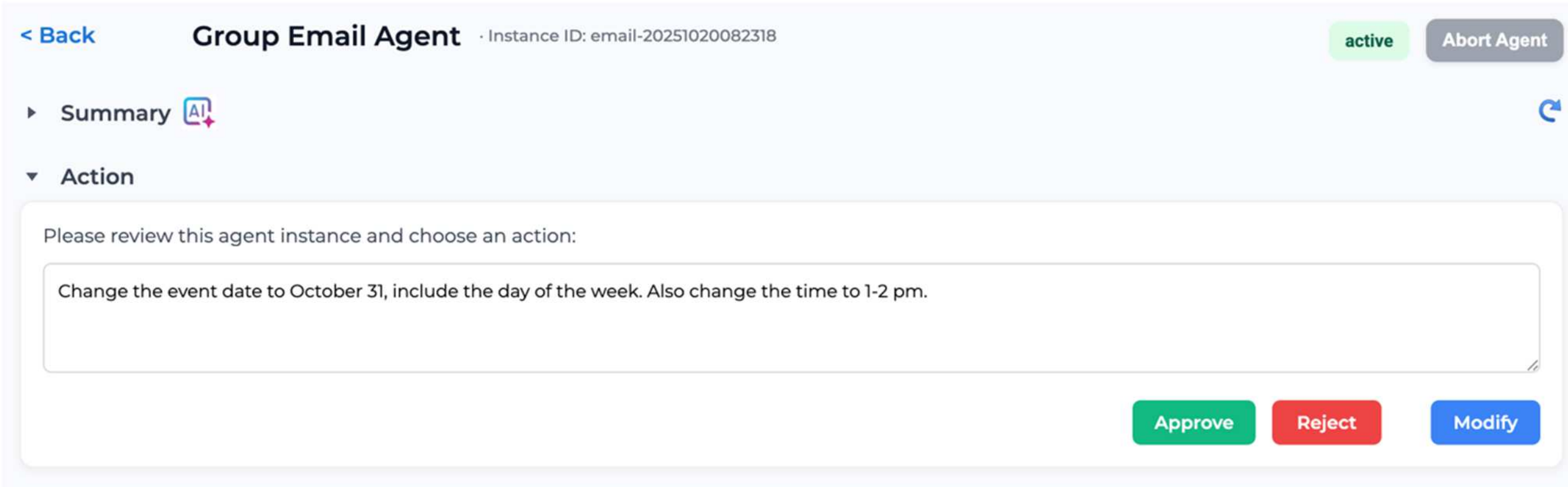

Fig. 3. Human-in-the-Loop approval interface within an agentic workflow.

### *C. Policy-Driven Governance and Escalation*

The proposed framework further supports executable governance policies that regulate agent behavior through operational constraints, escalation conditions, and HITL controls. Policies may define permitted actions, restricted operations, approval thresholds, audit requirements, and fallback behaviors across different execution contexts.

To support adaptive governance, the framework allows policies to combine deterministic rule evaluation with contextual LLM-based assessment. Deterministic rules may identify explicitly restricted behaviors, while LLM-based evaluators may assess ambiguous or situationally dependent actions requiring semantic interpretation. This hybrid approach enables operational policies to support both strict enforcement and contextual supervision within continuously operating agentic systems.

Table I illustrates representative policy controls used to govern autonomous actions, escalation conditions, and risk-sensitive operations within an enterprise agent environment.

TABLE I. EXAMPLE POLICY CONTROLS FOR GOVERNED AGENT AUTONOMY

| **Policy Category** | **Example Governance Rule** | **Control Type** |
|---|---|---|
| Sensitive Search Queries | Escalate searches containing credential or financial keywords | Hard |
| Sensitive File Access | Block reads of SSH keys, credential files, or system secrets | Hard |
| Destructive Shell Commands | Escalate commands containing destructive filesystem operations | Hard + LLM validation |
| Out-of-Scope File Writes | Review writes outside approved directories | Soft + LLM validation |
| Git Repository Modification | Require approval for force-push or history rewrite operations | Hard |
| Untrusted Email Recipients | Require approval for external recipients outside trusted lists | Hard |
| High-Value Payments | Escalate transactions exceeding policy thresholds | Hard |
| Volatile Stock Trading | Flag trading of high-risk securities for review | Soft |
| Human Override Logging | Record approvals, rejections, and intervention rationale | Auditability |

The policy framework supports both “hard” controls for mandatory escalation or blocking and “soft” controls for risk-sensitive or context-dependent evaluation. Policies may combine deterministic rules with contextual LLM-based assessment to support adaptive supervision.

Policy rules are represented using a structured schema in which each criterion specifies an identifier, applicable action or tool, rule type, evaluation method, and resulting governance action. Deterministic criteria use a restricted Python-compatible expression syntax over typed runtime attributes such as strings, numbers, Booleans, lists, and enumerated values. Hard rules represent mandatory constraints whose violation results in blocking or human escalation, while soft rules produce advisory or risk signals that may be combined with contextual evaluation before an action is allowed or escalated.

The following example illustrates a simplified governance policy used in a continuously operating autonomous agent environment.

```
{
  "policy_type": "eval",
  "hitl": {
    "required_for": ["high_risk_action"]
  },
  "criteria": [
    {
```

```
      "id": "sensitive_file_access",
      "tool": "read",
      "hardness": "hard",
      "rule": "path contains credential or key files",
      "action": "require_human_approval"
    },
    {
      "id": "destructive_shell_command",
      "tool": "exec",
      "hardness": "hard",
      "evaluation": "compute_then_llm",
      "action": "escalate"
    }
  ]
}
```

## VI. Conclusion and Future Work

This paper presented a Three-Pillar Governance Framework for Safe and Trustworthy AI Agents based on the 3P Model of transparency, accountability, and trustworthiness. The framework defines operational conditions for growing autonomy in continuously operating agentic systems through lifecycle observability, decision traceability, human-in-the-loop (HITL) oversight, and progressive validation of autonomous behavior. By integrating governance mechanisms directly into agent execution environments, the framework supports controlled progression toward higher levels of autonomy while maintaining operational safety and alignment.

Future work will extend the framework in two directions. First, operational signals collected through agent activity records, escalation events, and Human-in-the-Loop workflows can be leveraged to improve agent behavior over time. Beyond increasing the proportion of automated decisions, these signals may support improved decision quality, adaptive tool usage, and more effective execution strategies within dynamic operational environments.

Second, the scope of the 3P framework can be expanded from individual agents to coordinated multi-agent systems. By aggregating and analyzing activities across agents, future systems may support overseer layers that incorporate cross-agent context and shared operational memory to coordinate execution order, manage dependencies, allocate resources, and regulate collaborative workflows. Such orchestration capabilities may enable more structured and efficient multi-agent environments while reducing conflicts, redundant actions, and misalignment across autonomous agents.

The framework has not yet been validated through production deployment or comparative empirical evaluation. Its effectiveness also depends on appropriate policy design, risk thresholds, and human oversight. Future work should evaluate the framework across different agent types and operational environments.

## References

[1] T. Sanders, "How AI agents are overcoming market hype to deliver real business impact," 2025 AI Agents G2 Insight Report, October 2025. [Online]. Available: https://company.g2.com/news/2025-ai-agent-report

[2] F. Zanzotto, "Viewpoint: human-in-the-loop artificial intelligence," Journal of Artificial Intelligence Research 64, 2019, pp. 243-252.

[3] X. Wu, et al., "A survey of human-in-the-loop for machine learning," arXiv preprint arXiv:2108.00941, April 2022. [Online]. Available: https://arxiv.org/abs/2108.00941

[4] E. Mosqueria-Rey, E. Hernández-Pereira, D. Alonso-Ríos, J. Bobes-Bascarán, Á. Fernández-Leal, "Human-in-the-loop machine learning: a state of the art," Artificial Intelligence Review (2023) 56:3005–3054.

[5] B. Wixom, I. Someh, and R. Gregory, "AI alignment: a new management paradigm," MIT Center for Information Systems Research (MIT CISR), no. XX-11. November 2020.

[6] K. Burnham, "New framework helps companies build secure AI systems," MIT Management Sloan School, July 2025. [Online]. Available: https://mitsloan.mit.edu/ideas-made-to-matter/new-framework-helps-companies-build-secure-ai-systems

[7] F. Bellos, et al., "Towards effective human-in-the-loop assistive ai agents," arXiv preprint arXiv:2507.18374. [Online]. Available: https://arxiv.org/abs/2507.18374

[8] H. Mozannar, et al., "Magentic-UI: towards human-in-the-loop agentic systems," Microsoft Research AI Frontiers. arXiv:2507.22358 (v1). July 2025. [Online]. Available: https://arxiv.org/abs/2507.22358

[9] C. Retzlaff, et al., "Human-in-the-loop reinforcement learning: a survey and position on requirements, challenges, and opportunities," Journal of Artificial Intelligence Research 79, 2024, pp.359-415

[10] A. Ren, et al., "Robots that ask for help: uncertainty alignment for large language model planners," Google DeepMind. arXiv:2307.01928 (v2). September 2023, [Online]. Available: https://arxiv.org/abs/2307.01928

[11] A. Danielle, et al., "A roadmap for governing ai: technology governance and power sharing liberalism," ASH Center for Democratic Governance and Innovation. Harvard Kennedy School, January 2024.

[12] L. Barroso and P. Mello, "Artificial intelligence: promises, risks, and regulation: something new under the sun," CARR Center for Human Rights Policy. Harvard Kennedy School, December 2024.

[13] S. Natarajan, S. Mathur, S. Sidheekh, W. Stammer, and K. Kersting, "Human-in-the-loop or ai-in-the-loop? automate or collaborate?" The Thirty-Ninth AAAI Conference on Artificial Intelligence (AAAI-25), March 2025.

[14] E. Tabassi, "Artificial Intelligence Risk Management Framework (AI RMF 1.0)," National Institute of Standards and Technology, NIST AI 100-1, Jan. 2023, doi: 10.6028/NIST.AI.100-1.

[15] ISO/IEC, "ISO/IEC 42001:2023 Information technology—Artificial intelligence—Management system," International Organization for Standardization and International Electrotechnical Commission, Geneva, Switzerland, 2023.

[16] European Parliament and Council of the European Union, "Regulation (EU) 2024/1689 laying down harmonised rules on artificial intelligence (Artificial Intelligence Act)," Official Journal of the European Union, Regulation (EU) 2024/1689, Jul. 2024.

[17] OWASP Foundation, "OWASP Top 10 for LLM Applications 2025," OWASP GenAI Security Project, Nov. 2024. [Online]. Available: OWASP GenAI Security Project.

[18] J. Wang, L. Zhang, Y. Huang, and J. Zhao, "Safety of Autonomous Vehicles," Journal of Advanced Transportation, October 2020.

[19] M. Khan, et al., "Autonomous Driving—Are We There Yet?," A Review of Research Literature. ACM Journals, ACM Computing Surveys (CSUR), Vol. 55, Issue 2, Article No. 27, January 2022.

[20] E. Cheng and J. Cheng, "A decoupled human-in-the-loop system for controlled autonomy in agentic workflows," arXiv preprint arXiv:2604.23049, April 2026. [Online]. Available: https://arxiv.org/abs/2604.23049

[21] J. Cheng, "Context-Aware Prompt Enhancement (CAPE) Framework for a Multi-Agent Application System," AI Laboratory, Inquiryon, Inc., unpublished.